# A new Initial Centroid finding Method based on Dissimilarity Tree for K-means Algorithm

Abhishek Kumar[1], Suresh Chandra Gupta[2]
*Electrical Engineering Department*
*Indian Institute of Technology (B.H.U)*
*Varanasi India*
Email: [1]abhishek.kumar.eee13@iitbhu.ac.in, [2]scgupta.eee@iitbhu.ac.in

*Abstract*—Cluster analysis is one of the primary data analysis technique in data mining and K-means is one of the commonly used partitioning clustering algorithm. In K-means algorithm, resulting set of clusters depend on the choice of initial centroids. If we can find initial centroids which are coherent with the arrangement of data, the better set of clusters can be obtained. This paper proposes a method based on the Dissimilarity Tree to find, the better initial centroid as well as every bit more accurate cluster with less computational time. Theory analysis and experimental results indicate that the proposed method can effectively improve the accuracy of clusters and reduce the computational complexity of the K-means algorithm.

## I. INTRODUCTION

Clustering is a method of assembling similar data into a set of clusters. So, that the data objects within the clusters are similar, whereas object located to different clusters differ. Clustering is widely practiced in numerous arenas, for example image processing, machine learning, marketing, medicines, data compression, information retrieval and so on [1]. Clustering algorithms are basically split into two classes: Partitioning algorithms and Hierarchical algorithms. A Partitioning algorithms partition the dataset into a number of sets in a single step while a Hierarchical algorithms divide the given dataset into smaller subset in a hierarchical manner [2].

Numerous methods are available to solve clustering based problems [3]. K-means algorithm is a commonly used partitioning clustering algorithms applied in numerous domains, including the initialization of some computationally expensive algorithms like Gaussian mixtures, learning vector quantization, radial basis function and Markov's models etc. [4]. But k-means clustering algorithm selects initial centroids randomly and final cluster strongly based on the choice of initial centroids. Thus, it affects the computational time and accuracy of the cluster. Cluster result and computational time will be different for different centroid. Number of iterations needed while executing the K-means clustering algorithms and efficiency are also depends on the initial centroids [5]. Various methods have been proposed in various literature to enhance the efficiency and accuracy of K-means clustering algorithms.

In this paper, a dissimilarity tree based method is presented to find the initial centroid and enhance the accuracy and efficiency of the K-means algorithm. This study is motivated by [6] in which a TP (Tree Pruning) algorithm is presented to tackle the problem of the initial medoids of PAM algorithm. This paper is structured as follows:

Segment 2: It gives the overview of related studies.
Segment 3: It discusses the standard K-means algorithm.
Segment 4: It introduced the proposed algorithm.
Segment 5: It experiment demonstrates the implement-
-ation of the modified algorithm.
Segment 6: It describes the conclusion and future work.

## II. RELATED WORK

The standard K-means algorithm is really sensitive to initial centroid [5]. Several methods have been proposed for finding the better initial centroid [7] [8] [9]. Some methods were also aimed to amend both the efficiency and accuracy of K-means clustering technique [10].

A. M. Fahim et al. [7] proposed an algorithm that require less execution time compared to K-means clustering technique. In [7], the author proposed to maintain the distance to the closest cluster of previous iteration and use it to compare with distance from new centroid in the next iteration. When the current distance is smaller than or equal to the previous one, the data object remains in its cluster and there is no requirement of computing again its distances from the other cluster centroids. This saved the computational time, but initial centroids are still selected randomly.

A. Bhattacharya et al. [8] proposed a advanced clustering algorithm, called "Divisive Correlation Clustering Algorithm (DCCA)" for clustering of genes. This algorithm is capable to generate clusters of datasets without using any initial centroids. The time complexity of this algorithm is very high.

Fang Yuan et al. [9] presented a method to find the initial centroid. This method generates more stable clusters compared to standards K-means algorithm. In this method the initial centroids are calculated, systematic way.

Md. Sohrab Mahmud et al. [10] proposed an algorithm to find the initial centroid. This method finds a weighted average score of the dataset, then sorted this score of all data objects and separated into k subset. Finally the average for each subset is taken as centroid. This algorithm generates good clusters in a minimum amount of running time.

## III. K-MEAN ALGORITHM

K-means is a partitioning type clustering algorithm used in data-mining and it is one of the most popular, simple and unsupervised learning algorithm [11]. The basic concept of this algorithm is to be clustered the given datasets D into k number of disjoint clusters. The algorithm consists two basic steps [5]. The first step is to select the K-initial centroids for each cluster randomly. Where k is the number of clusters. The second step is to take all data objects of dataset to the nearby centroids [5]. Euclidean distance largely used to calculate the length between all data objects and the centroids. When each of the data objects are inserted in some clusters the initial grouping is done. After this, the centroid of all clusters is again estimated by taking the average value of all the data objects of all clusters. Some data objects may update their cluster to other cluster. Again, we calculate new centroids and assigning data objects to the nearby centroid. This procedure goes on yet the convergence criteria have not been satisfied or the centroids have not become similar for two consecutive iterations [5]. The computational time complexity of the k-means algorithm is O (nkt). Where n is the total number of all data objects, k is the total number of clusters, t is the total no of iterations of the algorithm. Pseudo code for the K-means clustering algorithm is described below [10].

1) Euclidean distance $d(a_i,b_i)$ can be calculated as follows :

$$d(a_i,b_i) = \left[ \sum_{i=1}^{n} (a_i - b_i)^2 \right]^{1/2}$$

Where, $a_i$ is $(a_1, a_2......a_i)$

$b_i$ is $(b_1, b_2......b_n)$

2) Criterion function is set as follows:

$$E = \sum_{i=1}^{k} \sum_{x \in c_i} |x - x_i|^2$$

Where x is the target object

$x_i$ is the mean cluster $c_i$

The algorithm is as follows:

Input : O = {$O_1$, $O_2$, $O_3$ ....... $O_n$}

K = number of required clusters.

Output : C = A set of K desired clusters

Steps:
1. Randomly opt k data objects for initial centroids from O;
2. Assign data objects to their nearby cluster, which has nearest centroids;
3. Estimate the new intermediate value for all cluster;
4. Go to step-2 yet convergence criteria have not been satisfied.

## IV. PROPOSED ALGORITHM

The Dissimilarity Tree algorithm uses the mixed variable type formula to estimate the dissimilarity between data objects. Dissimilarity between objects a and b is defined as:

$$dist[a,b] = \frac{\sqrt{\sum_{f=1}^{m} (\delta_{ab}^{(f)} d_{ab}^{(f)})^2}}{\sum_{f=1}^{m} \delta_{ab}^{(f)}}$$

Where $\delta_{ab}^{f} = 0$ if either $x_f$ or $y_f$ is missing or $x_f = y_f = 0$; otherwise, $\delta_{xy}^{f} = 1$.

$d_{xy}^{(f)}$ is defined as:

$$d_{ab}^{(f)} = \frac{|a_f - b_f|}{\max_h h_f - \min_h h_f}$$

The steps of dissimilarity tree based algorithm borrow the steps of TP (Tree Pruning) algorithm from [6]. This algorithm contains four steps:

I. Estimate matrix D by using above formula.

II. Construct a MST t, by using matrix D.

III. Prune the (k-1) branches which have maximum weight. As a result, k- sub-trees will form.

IV. Calculate the initial centroid by computing intermediate value of data objects of k sub-trees.

The next example can described above steps:

Table 1: Data Table of 3- dimension sample object

| Objects | X | Y | Z |
|---------|---|---|---|
| 1. | 2 | 5 | 6 |
| 2. | 7 | 1 | 2 |
| 3. | 3 | 6 | 4 |
| 4. | 1 | 8 | 0 |
| 5. | 1 | 9 | 2 |
| 6. | 5 | 2 | 6 |
| 7. | 8 | 2 | 3 |
| 8. | 4 | 6 | 1 |
| 9. | 6 | 4 | 5 |
| 10. | 9 | 3 | 7 |

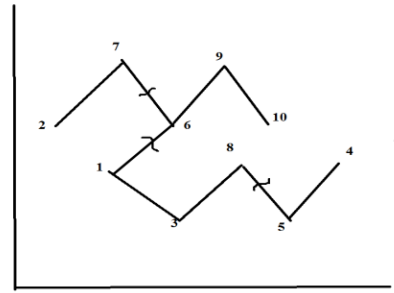

Figure 1: Minimum dissimilarity tree.

There are ten sample objects in the above case and these sample objects have to split into four clusters. Every sample object consists three ascriptions. Assigned data of sample objects is given in table l. We calculate test -1 ascription to find dissimilarity matrix $D_1$ and calculate test-2 ascription to find dissimilarity matrix $D_2$, also calculate test-3 ascription to find dissimilarity matrix $D_3$. Afterward that, a mean dissimilarity matrix $D_m$ can get by formula 1. The result are as follows:

$$D_1 = \begin{bmatrix} 0 & 0.625 & 0.125 & 0.125 & 0.125 & 0.375 & 0.75 & 0.25 & 0.5 & 0.875 \\ 0.625 & 0 & 0.5 & 0.75 & 0.75 & 0.25 & 0.125 & 0.375 & 0.125 & 0.25 \\ 0.125 & 0.5 & 0 & 0.25 & 0.25 & 0.25 & 0.5 & 0.125 & 0.375 & 0.75 \\ 0.125 & 0.75 & 0.25 & 0 & 0 & 0.5 & 0.875 & 0.375 & 0.625 & 1 \\ 0.125 & 0.75 & 0.25 & 0 & 0 & 0.5 & 0.875 & 0.375 & 0.625 & 1 \\ 0.375 & 0.25 & 0.25 & 0.5 & 0.5 & 0 & 0.375 & 0.125 & 0.125 & 0.5 \\ 0.75 & 0.125 & 0.5 & 0.875 & 0.875 & 0.375 & 0 & 0.5 & 0.25 & 0.125 \\ 0.25 & 0.375 & 0.125 & 0.375 & 0.375 & 0.125 & 0.5 & 0 & 0.25 & 0.625 \\ 0.5 & 0.125 & 0.375 & 0.625 & 0.625 & 0.125 & 0.25 & 0.25 & 0 & 0.375 \\ 0.875 & 0.25 & 0.75 & 1 & 1 & 0.5 & 0.125 & 0.625 & 0.375 & 0 \end{bmatrix}$$

$$D_2 = \begin{bmatrix} 0 & 0.5 & 0.125 & 0.375 & 0.5 & 0.375 & 0.375 & 0.125 & 0.125 & 0.25 \\ 0.5 & 0 & 0.625 & 0.875 & 1 & 0.125 & 0.125 & 0.625 & 0.375 & 0.25 \\ 0.125 & 0.625 & 0 & 0.25 & 0.375 & 0.5 & 0.5 & 0 & 0.25 & 0.375 \\ 0.375 & 0.875 & 0.25 & 0 & 0.125 & 0.75 & 0.75 & 0.25 & 0.5 & 0.625 \\ 0.5 & 1 & 0.375 & 0.125 & 0 & 0.875 & 0.875 & 0.375 & 0.625 & 0.25 \\ 0.375 & 0.125 & 0.5 & 0.75 & 0.875 & 0 & 0 & 0.5 & 0.25 & 0.125 \\ 0.375 & 0.125 & 0.5 & 0.75 & 0.875 & 0 & 0 & 0.5 & 0.25 & .125 \\ 0.125 & 0.625 & 0 & 0.25 & 0.375 & 0.5 & 0.5 & 0 & 0.25 & 0.375 \\ 0.125 & 0.375 & 0.25 & 0.5 & 0.625 & 0.25 & 0.25 & 0.25 & 0 & 0.125 \\ 0.25 & 0.25 & 0.375 & 0.625 & 0.25 & 0.125 & .125 & 0.375 & 0.125 & 0 \end{bmatrix}$$

$$D_3 = \begin{bmatrix} 0 & 0.5 & 0.25 & 0.75 & 0.5 & 0 & 0.375 & 0.625 & 0.125 & 0.125 \\ 0.5 & 0 & 0.25 & 0.25 & 0 & 0.5 & 0.125 & 0.125 & 0.375 & 0.625 \\ 0.25 & 0.25 & 0 & 0.5 & 0.25 & 0.25 & 0.125 & 0.125 & 0.125 & 0.375 \\ 0.75 & 0.25 & 0.5 & 0 & 0.5 & 0.75 & 0.25 & 0.125 & 0.625 & 0.875 \\ 0.5 & 0 & 0.25 & 0.5 & 0 & 0.5 & 0.125 & 0.125 & 0.375 & 0.625 \\ 0 & 0.5 & 0.25 & 0.75 & 0.5 & 0 & 0.375 & 0.625 & 0.125 & 0.125 \\ 0.375 & 0.125 & 0.125 & 0.25 & 0.125 & 0.375 & 0 & 0.25 & 0.25 & 0.5 \\ 0.625 & 0.125 & 0.375 & 0.125 & 0.125 & 0.625 & 0.25 & 0 & 0.5 & 0.75 \\ 0.125 & 0.375 & 0.125 & 0.625 & 0.375 & 0.125 & 0.25 & 0.5 & 0 & 0.25 \\ 0.125 & 0.625 & 0.375 & 0.875 & 0.625 & 0.125 & 0.5 & 0.75 & 0.25 & 0 \end{bmatrix}$$

$$D_m = \begin{bmatrix} 0 & 0.541 & 0.167 & 0.417 & 0.375 & 0.75 & 0.5 & 0.333 & 0.25 & 0.412 \\ 0.541 & 0 & 0.625 & 0.583 & 0.292 & 0.125 & 0.125 & 0.375 & 0.292 & 0.375 \\ 0.167 & 0.625 & 0 & 0.333 & 0.292 & 0.333 & 0.35 & 0.167 & 0.25 & 0.5 \\ 0.417 & 0.583 & 0.333 & 0 & 0.125 & 0.667 & 0.667 & 0.25 & 0.583 & 0.833 \\ 0.375 & 0.292 & 0.292 & 0.125 & 0 & 0.625 & 0.625 & 0.292 & 0.542 & 0.792 \\ 0.75 & 0.125 & 0.333 & 0.667 & 0.625 & 0 & 0.25 & 0.417 & 0.167 & 0.25 \\ 0.5 & 0.125 & 0.35 & 0.667 & 0.625 & 0.25 & 0 & 0.417 & 0.25 & 0.25 \\ 0.333 & 0.375 & 0.167 & 0.25 & 0.292 & 0.417 & 0.417 & 0 & 0.333 & 0.583 \\ 0.25 & 0.292 & 0.25 & 0.583 & 0.542 & 0.167 & 0.25 & 0.333 & 0 & 0.25 \\ 0.412 & 0.375 & 0.5 & 0.833 & 0.792 & 0.25 & 0.25 & 0.583 & 0.25 & 0 \end{bmatrix}$$

Look for the dissimilarity $D_m$ to draw a minimal spanning tree called Dissimilarity tree. Ten nodes and the nine branches build a Dissimilarity tree as fig 1 shows. It is required to take rid of the branch dist-(6,7), dist-(1,6), dist-(4,8) in order to cut the tree into four sub-tree, which were $c_1 = \{1,3,8\}$, $c_2 = \{2,7\}$, $c_3 = \{6,9,10\}$, $c_4 = \{5,4\}$. These four sub-trees are clusters. From these, we are easily calculate initial centroids.

The above described method to find initial centroids of the clusters are nearer to original centroids than the K-means algorithm where centroids are picked up arbitrary. When k-means algorithm use this technique to find initial centroid, it converge faster than the standard K-means algorithm.

## V. EXPERIMENTAL RESULTS

The experimental environment of this context was AMD CPU, 2 G EMS Memory, 500 GB hard disk and Windows 8.1 OS. MATLAB 2013a is used as tool to validate the validity of improved algorithm. This paper select the multivariate datasets, selected from the UCI repository of machine learning, databases [12], that is utilized to test both the efficiency and accuracy of the proposed K-means algorithm. The same datasets have taken as input to the standard algorithm and modified algorithm. Experiment compares the accuracy and the entire running time of the sets of clusters of both the algorithms. Both algorithms requires number of desired clusters as an input. Standard K-means algorithm will take initial centroid randomly but modified algorithm will compute initial centroid automatically and find optimal centroids by the program. This paper uses wine [12], iris [12] as the datasets and table 2: shows some features of datasets.

Table 2: Characteristics of datasets.

| Datasets | No of attributes | No of records |
|---|---|---|
| Wine | 15 | 178 |
| Iris | 4 | 150 |

Each algorithm is executed ten times for both datasets. In each experiment the accuracy and timelines were computed and taken the mean values of all experiments. Tabled 3 shows the performance comparison of the algorithms. The results also showed with help of bar charts in the fig 2 & 3.

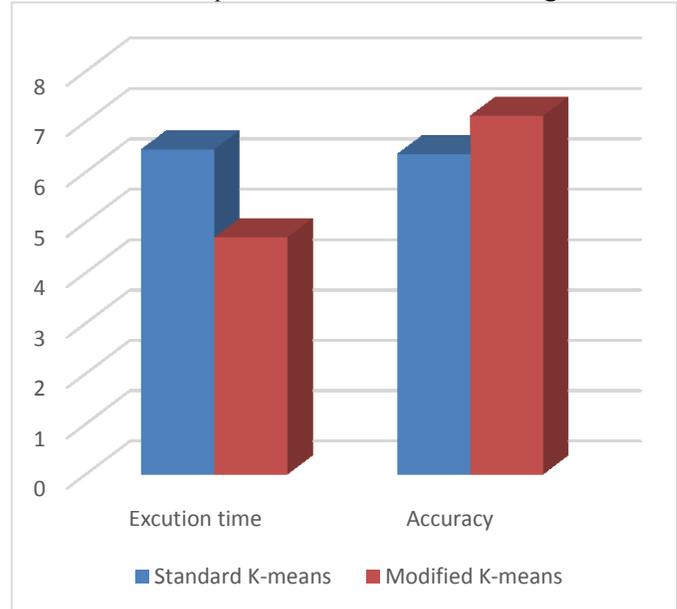

Fig 2: Performance comparison chart for Wine.

Table 3: Clustering result.

| Dataset | No of cluster | Algorithm | Execution time (sec) | Accuracy (%) |
|---------|---------------|-----------|---------------------|--------------|
| Wine | 3 | Standard K-means | 0.0644 | 63.5 |
| Wine | 3 | Modified K-means | 0.0470 | 71.1 |
| Iris | 3 | Standard K-means | 0.0505 | 75.3 |
| Iris | 3 | Modified K-means | 0.0250 | 80.1 |

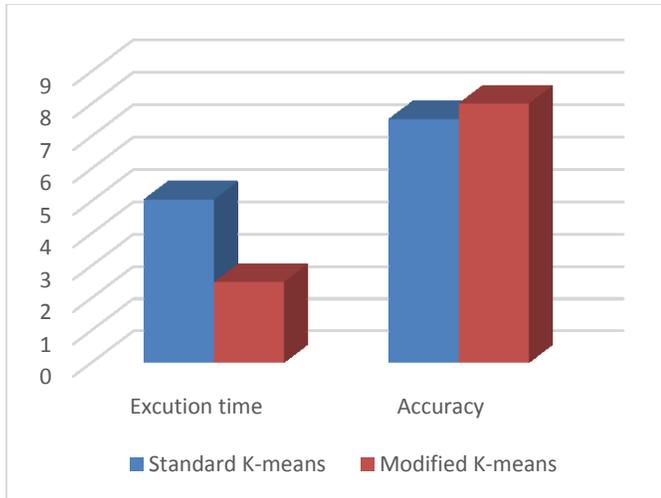

Fig 3: Performance comparison chart for Iris.

The outcomes of experiment show that the formation of clusters has led a less measure of execution time and quality of clusters are better in the modified algorithm as compared to the standard algorithm.

## VI. CONCLUSION AND FUTURE WORK

K-means algorithm is widely used in many application area but in this method the quality of the final clusters are |highly depend on the initial centroid, which are selected randomly. The modified algorithm is generating cluster result more accurately and efficiently compared to the standard K-means algorithm. This paper presents a dissimilarity tree algorithm which computes the initial centroids that reduce the number of iteration in K-means algorithm. One limitation of this algorithm is that it still use the desired no of clusters as a input. Automatic determination of the number of cluster according to the distribution of data objects in datasets is suggested as a future work

## REFERENCES

[1] Sun Shibao, Qin Keyun, "Research on Modified k-means Data Cluster Algorithm" I. S. Jacobs and C. P. Bean, "Fine particles, thin films andex change anisotropy," Computer Engineering, vol.33, No.13, pp.200–201,July 2007.

[2] Margret H. Dunham, *Data Mining-Introductory and Advanced Concepts*, Pearson Education, 2006.

[3] Madhu Yedla, Srinivasa Rao Pathakota, T M Srinivasa, "Enhancing K-means Clustering Algorithm with Improved Initial Center," International Journal of Computer Science and Information Technologies (IJCSIT), vol. 1(2), 2010, 121-125.

[4] Leon Bottou, Neuristique, Yoshua Bengio, " Convergence Properties of the K-means Algorithms," Dept, IRO, IESI-CNR.

[5] J. Han and M. Kamber, *Data Mining Concepts and Technique,* Morgan Kaufmann Publishers, San Diego, 2001.

[6] Feng Bo, Hao Wenning, Chen Gang, Jin Dawei, Zhao Shuining, " An Improved PAM Algorithm for Optimizing Initial Cluster Centre," IEEE, 2012, 978-1-4673-2008-5/12.

[7] A. M. Fahim, A. M. Salem, F. A. Torkey, M. A. Ranadan, " An Efficient Enhanced K-means Clustering Algorithm," Journel of Zejiang University, 10(7): 16261633, 2006.

[8] A. Bhattacharya, R. K. De, "Divisive Correlation Clustering Algorithm (DCCA) for Grouping of Genes: detecting varying patterns in expression profiles," Bioinformatics, vol. 24, pp. 1359-1366, 2008.

[9] F. Yuan, Z. H. Meng, H. X. Zhangz, C. R. Dong, "A New Algorithm to Get the Initial Centroids," proceedings of the 3rd International Conference on Machine Learning and Cypernetics, pp. 26-29, August 2004.

[10] S. Mahmud, M. Rahman, N. Akhtar, " Improvement of K-means Clustering Algorithm with Better Initial Centroids based on Weighted Average," 7th International Conference on Electrical and Computer Engineering, Dhaka, PP. 647-650, 20-222 December 2012,

[11] Sun Jigui, Liu Jie, Zhao Lianyu, "Clustering Algorithm Research," Journal of software, vol. 19, no 1, pp. 48-61, January 2008.

[12] The UCI Repository website (2014). [online]. Available: http://archive.ics.uci.edu/